%% file: main.tex
\pdfoutput=1
\documentclass[11pt]{article}

\usepackage[final]{acl}

\usepackage{times}
\usepackage{latexsym}

\usepackage[T1]{fontenc}
\usepackage[utf8]{inputenc}

\usepackage{microtype}

\usepackage{inconsolata}

\usepackage{graphicx}

\usepackage{algorithm}
\usepackage{algorithmic}
\usepackage{amsmath}
\usepackage{booktabs}
\usepackage{multirow}
\usepackage{graphicx}
\usepackage{tabularx} 
\usepackage{arydshln}
\usepackage[russian,english]{babel}

\input{macros}
\title{ConLID: Supervised Contrastive Learning for \\Low-Resource Language Identification}

\author{
Negar Foroutan\textsuperscript{1}\thanks{Equal contribution},
Jakhongir Saydaliev\textsuperscript{1}\footnotemark[1],
Grace Kim\textsuperscript{2},
Antoine Bosselut\textsuperscript{1} \\
\textsuperscript{1}EPFL \quad
\textsuperscript{2}The University of Texas at Austin \\
  \small{
   \textbf{Correspondence:} \href{mailto:antoine.bosselut@epfl.ch}{\{negar.foroutan, antoine.bosselut\}@epfl.ch}
   }
}

\begin{document}
\maketitle
\begin{abstract}
Language identification (LID) is a critical step in curating multilingual LLM pretraining corpora from web crawls. While many studies on LID model training focus on collecting diverse training data to improve performance, low-resource languages -- often limited to single-domain data, such as the Bible -- continue to perform poorly. To resolve these imbalance and bias issues, we propose a novel supervised contrastive learning (SCL) approach to learn domain-invariant representations for low-resource languages. We show that our approach improves LID performance on out-of-domain data for low-resource languages by 3.2 percentage points, while maintaining its performance for the high-resource languages.\footnote{Our code and model are available at \href{https://github.com/epfl-nlp/ConLID}{https://github.com/epfl-nlp/ConLID}.}
\end{abstract}

\input{sections/introduction}

\input{sections/related_work}
\input{sections/methodology}
\input{sections/experimental_setup}
\input{sections/analysis}

\input{sections/conclusion}
\input{sections/limitations}

\section*{Acknowledgments}
We thank Deniz Bayazit, Angelika Romanou, and Ayush Kumar Tarun for their helpful discussions and feedback on our paper. 
We also gratefully acknowledge the support of the Swiss National Science Foundation (No. 215390), the European Research Council (Starting grant no. 101222478, RESPECT-LM), the AI2050 program at Schmidt Sciences (Grant \#G-25-69783), Sony Group Corporation, and the Swiss National Supercomputing Center (CSCS) in the form of an infrastructure engineering and development project.

\bibliography{custom}

\appendix
\input{sections/appendix_hyperparameters}

\end{document}

%% file: macros.tex
\newcommand{\ie}{\textit{i.e.}}
\newcommand{\eg}{\textit{e.g.}}
\newcommand{\our}{ConLID}
\newcommand{\glotlidm}{GlotLID-M}
\newcommand{\MS}{\text{\our-S}}%_\text{S}}}$}
\newcommand{\MSB}{\text{\our-S-B}}%_\text{S}}}$}
\newcommand{\MH}{\text{\our-H}}%_\text{S}}}$}
\newcommand{\CE}{$\text{LID}_{\text{CE}}$}%_\text{S}}}$}
\newcommand{\CEB}{$\text{LID}_{\text{CE}}-\text{B}$}%_\text{S}}}$}
\newcommand{\SCL}{$\text{LID}_{\text{SCL}}$}%_\text{S}}}$}

\newcommand{\MSensemble}{\MS+LID\textsubscript{CE}}
\newcommand{\MGensemble}{\glotlidm+\MS}

\newcommand{\flores}{FLORES-200}
\newcommand{\udhr}{UDHR}
\newcommand{\glotlidc}{\texttt{GlotLID-C}}
\newcommand{\glotlidctest}{\texttt{GlotLID-C-test}}
\newcommand{\glotlidctrain}{\texttt{GlotLID-C-train-down}}
\newcommand{\finewebtwo}{FineWeb-2}

\definecolor{darkgreen2}{HTML}{009B55}
\definecolor{myblue2}{HTML}{29abe2}
\definecolor{myorange2}{HTML}{f7931e}

\definecolor{mypurple2}{HTML}{9823FF}

\newif\ifcomments
\commentstrue
\ifcomments
    \usepackage[normalem]{ulem}
    \definecolor{ABpurple}{rgb}{0.8,0.0,0.8}
    \newcommand\abi[1]{\textcolor{ABpurple}{#1}} %inline
    \newcommand\abm[1]{{\marginparwidth=2cm \marginpar{\raggedright\tiny\textcolor{ABpurple}{\textsf{{\bfseries AB\@:} #1}}}}} %margin comment
    \newcommand\abs{\bgroup\markoverwith{\textcolor{ABpurple}{\rule[.4ex]{2pt}{0.8pt}}}\ULon} %crossout
\else
    \newcommand\ab[1]{}
    \newcommand\abi[1]{\ignorespaces}
    \newcommand\abm[1]{}
    \newcommand\abs[1]{#1}
\fi

\definecolor{myred}{HTML}{E11D3F}
\definecolor{myorange}{HTML}{F04A00}
\definecolor{myblue}{HTML}{1974D2}
\definecolor{mypurple}{HTML}{6A0DAD}
\ifcomments

    \newcommand{\todo}[1]{{\scriptsize\color{blue!80!black}[{\bf TODO: }\textsf{#1}]}}
    
\else
    \excludecomment{todoenv}
    \excludecomment{noteenv}
    \newcommand{\todo}[1]{}
\fi

%% file: sections/introduction.tex
\section{Introduction}
\label{sec:introduction}

Language identification (LID) is a fundamental preprocessing task in natural language processing (NLP). In recent years, LID has been a crucial step in the creation of large-scale textual pretraining corpora. LID models are used to label heterogeneous, multilingual document sources in web crawls. Subsequently, these labels are used to either filter non-target language content \citep{rae2022scalinglanguagemodelsmethods, dubey2024llama3herdmodels, li2024datacomplmsearchgenerationtraining} or leverage language identification for effective data scheduling in multilingual pretraining \citep{Conneau2019UnsupervisedCR, degibert2024newmassivemultilingualdataset, imanigooghari-etal-2023-glot500,laurenon2023bigscience,nguyen2023culturax}.

\begin{figure}[ht]
    \centering
    \includegraphics[width=0.48\textwidth]{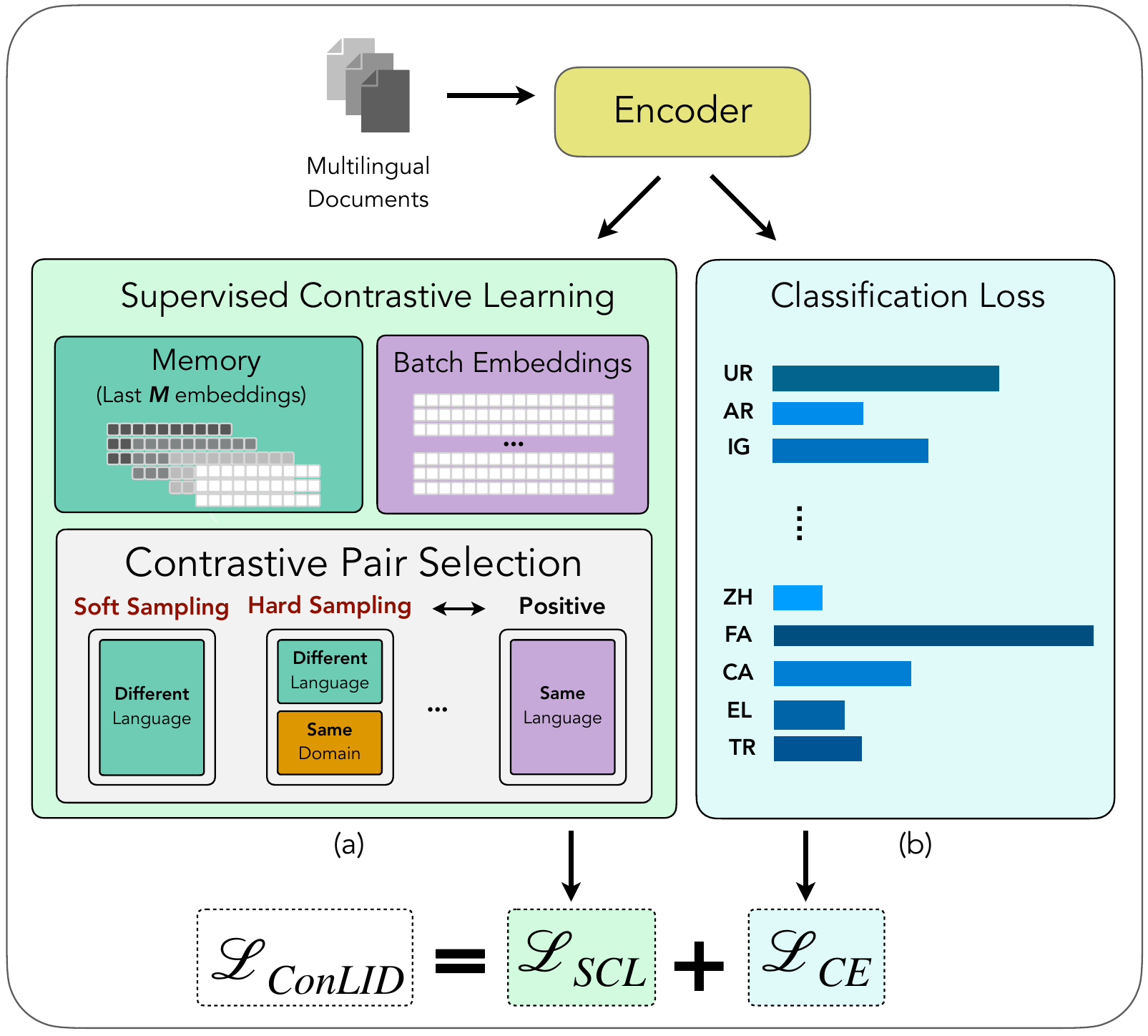}
    \caption{Overview of training ConLID. Each sentence is first processed by an encoder that generates a sentence representation. This representation is then passed to two components: (a) a Supervised Contrastive Learning (SCL) module, and (b) a feed-forward neural network that serves as the classification head. The final loss used for training is a combination of the classification loss and the SCL loss.}

    \label{fig:enter-label}
\end{figure}

The application of these LID systems to web crawls requires scalable methods that can be efficiently applied to large document collections \citep{penedo2024fineweb2}. Consequently, current approaches to LID \citep{kargaran-etal-2023-glotlid, nllbteam2022languageleftbehindscaling}
predominantly use simple models~\citep{joulin-etal-2017-bag} trained with cross-entropy (CE) loss.
While these supervised approaches have been sufficient to practically solve LID for high-resource standard languages \citep{mcnamee2005solved}, low-resource languages and dialects remain difficult to identify and categorize accurately for two reasons. First, data in these languages is often scarce or even mislabeled \citep{roy_information_2022,yu_beyond_2022,joshi_state_2020}, leading to class imbalances during training \citep{kargaran-etal-2023-glotlid}. Second, available data is often concentrated in specific domains, such as religious texts (\ie, Bible translations; \citealp{agic-vulic-2019-jw300,imanigooghari-etal-2023-glot500,ebrahimi-kann-2021-adapt}). This domain entanglement yields models trained on narrow datasets that often fail to generalize effectively to diverse types of text.

To mitigate these limitations, we propose a novel approach that employs supervised contrastive learning (SCL; \citealp{NEURIPS2020_d89a66c7}) for LID. 
Our SCL objective explicitly encourages representations of texts from the same language to cluster together while pushing those from different languages further apart in the embedding space. Combined with a classic CE loss, our dual-objective framework promotes learning less biased language representations that are robust to in-language domain shifts, enabling better generalization across diverse text types. A critical component of our SCL training method is a \textit{memory bank} \cite{tan-etal-2022-domain} from which we scalably sample more diverse positive and negative examples, effectively increasing our training batch size while keeping memory requirements constant. This augmentation enhances our contrastive learning approach, yielding stable training and discriminative language representations. To mitigate the second challenge, we employ a novel \textit{hard negative} mining scheme for SCL training, where data sources in our training set are labeled with {domain} information and negative samples are drawn from different languages in the same domain to learn language-specific, domain-invariant representations during training.

We evaluate our approach on three benchmark datasets: \glotlidc{} \citep{kargaran-etal-2023-glotlid}, \flores{} \cite{nllbteam2022languageleftbehindscaling}, and \udhr,\footnote{\url{https://huggingface.co/datasets/cis-lmu/udhr-lid}} and demonstrate that our approach leads to improvements over state-of-the-art LID methods, particularly in out-of-domain evaluations. Specifically, our best SCL-based approaches achieve a 3.2 percentage points improvement over CE-based models for low-resource languages, and a 5.4 percentage points improvement for languages with data from diverse domains, indicating enhanced domain generalization. 

Moreover, we conduct an in-depth analysis of underperforming languages in out-of-domain evaluations and observe that misclassifications predominantly occur among linguistically related languages, highlighting potential challenges in LID for closely related language pairs.
Finally, we assess our approach in a real-world scenario by evaluating it on \finewebtwo{} \citep{penedo2024fineweb2}, a large-scale multilingual pretraining corpus. We compare its performance against state-of-the-art LID systems for low-resource languages, providing insights into the practical significance of LID systems.

Our key contributions are as follows: We introduce the first application of supervised contrastive learning (SCL) for domain generalization in LID. We provide a comprehensive analysis of misclassified languages in out-of-domain settings, shedding light on their linguistic and data-related characteristics.
These contributions advance the robustness of language identification, particularly for low-resource and domain-diverse scenarios, paving the way for more reliable multilingual NLP systems.

%% file: sections/related_work.tex
\section{Related Work}
\label{sec:related_works}
\subsection{Language Identification}
LID tools are often used to collect multilingual corpora \cite{Dunn2020, penedo2024fineweb2, kargaran-etal-2023-glotlid}, and there have been extensive work on developing LID tools focusing on covering as diverse languages as possible \cite{kargaran-etal-2023-glotlid, adebara-etal-2022-afrolid, dunn-edwards-brown-2024-geographically, burchell-etal-2023-open, nllbteam2022languageleftbehindscaling}.
While many of LID models' architectures are based on FastText \cite{kargaran-etal-2023-glotlid, dunn-edwards-brown-2024-geographically, burchell-etal-2023-open, nllbteam2022languageleftbehindscaling}, which fails to learn domain-invariant representation of languages, some models are based on Transformer \cite{adebara-etal-2022-afrolid, caswell-etal-2020-language, sindane-marivate-2024-n}, which is less efficient for real-world scenarios.

\subsection{Contrastive Learning}
Contrastive Learning (CL) has been widely used in Computer Vision for learning generalizable representations in a self-supervised setting, without relying on labels \cite{10.5555/3524938.3525087}. Supervised Contrastive Learning (SCL) extends this idea to leverage labeled data, grouping same-class examples while separating different-class ones \cite{NEURIPS2020_d89a66c7}. It is shown that the performance of SCL is largely affected by the batch size, as the contrasting examples are selected within the batch to compute the contrastive loss \cite{NEURIPS2020_d89a66c7, gunel2021supervised, tan-etal-2022-domain}.

In NLP, while the previous works have applied SCL for in-domain \cite{gunel2021supervised} and out-of-domain \citep{tan-etal-2022-domain, khondaker-etal-2023-cross} sentence-level classification tasks, those works (1) focus on the cases with only a few class labels (typically, less than 10), which allows for a good use of SCL, and (2) using Transformer-based models. To the best of our knowledge, we are the first to apply SCL to the LID task with significantly larger label classes (\textasciitilde2,000) and using a simple linear classifier instead of a Transformer-based model.

%% file: sections/methodology.tex
\section{Methodology}
\label{sec:method}

In the following, we provide a detailed description of our base model architecture (\S\ref{sec:model_description}),
the principles of supervised contrastive learning (SCL; \S\ref{sec:scl}), our positive/negative pair selection approach (\S\ref{sec:negative_pair_selection}), and an optimized training strategy for the SCL objective, particularly suited for scenarios involving a large number of classes (\S\ref{sec:memory_bank}).

\subsection{Base Classifier}
\label{sec:model_description}
We train the base model following the methodology employed by FastText \citep{joulin-etal-2017-bag}. Given an input sentence, each word in the sentence is decomposed into character-level n-grams. The sentence representation is, then, obtained by computing the average of the vectors corresponding to these n-grams, along with the word-level embedding. The training process proceeds as follows: 

Initially, the training dataset is partitioned into mini-batches. For a sampled mini-batch of size $B$, we define $S = \left\{ x_i, y_i \right\}_{i=1}^{B}$, where $x_i$ represents the input text, and $y_i$ denotes its corresponding language class.
Each input text $x_i$ is represented by an embedding, $\mathbf{z}_i$, using an encoder function $\mathbf{z}_i = E(x_i)$, where $E$ corresponds to the encoder.
The resulting embeddings $\mathbf{z}_i$ are subsequently passed through a feed-forward neural network $f$, which serves as the classification head.
The model is trained using the cross-entropy (CE) loss, formulated as:
\begin{equation}
    \mathcal{L}_{CE} = -\frac{1}{B} \sum_{i=1}^{B} y_i \log \big( f(\mathbf{z}_i) \big)
    \tag{1}
    \label{eq:cross_entropy}
\end{equation}

\subsection{Supervised Contrastive Learning}
\label{sec:scl}
The Supervised Contrastive Learning (SCL) objective minimizes the distance between representations of positive pairs (of the same language) while maximizing the separation between negative pairs (of different languages) to enhance the discriminative capability of learned hidden representations.
Formally, given a mini-batch $S$ of size $B$, the SCL objective is computed as follows:
\begin{equation}
    \begin{split}
    \mathcal{L}_{SCL} = -\frac{1}{B} \sum_{i=1}^{B} \Bigg(\log \Big(\sum\limits_{\mathbf{z}_p \in P(i)} e^{\mathbf{z}_i \cdot \mathbf{z}_p / \tau}\Big) -  \\ \log\Big({\sum\limits_{\mathbf{z}_p \in P(i)} e^{\mathbf{z}_i \cdot \mathbf{z}_p / \tau} + \sum\limits_{\mathbf{z}_n \in N(i)} e^{\mathbf{z}_i \cdot \mathbf{z}_n / \tau} }\Big)\Bigg)
    \end{split}
    \tag{2}
    \label{eq:contrastive_loss}
\end{equation}

\noindent where $\mathbf{z}_i$ corresponds to the representation of a single data sample, $\mathbf{z}_p$ are positive samples drawn from a positive pair set $P(i) \equiv \left\{z_j \in S, y_j=y_i \right\}$ (\ie, all samples in $S$ that share the same label as $\mathbf{z}_i$), $\mathbf{z}_n$ are negative samples drawn from the negative pair set $N(i)$ (defined according to \S\ref{sec:negative_pair_selection}), and $\tau$ is the temperature.
The final combined loss function is then:

\begin{equation}
    \mathcal{L} = \mathcal{L}_{CE} + \mathcal{L}_{SCL}
    \tag{3}
    \label{eq:both}
\end{equation}

\subsection{Negative Pairs Selection}
\label{sec:negative_pair_selection}
We select negative pairs for training (Eq. \ref{eq:contrastive_loss}) using two strategies:
\paragraph{Soft selection.} Given a batch $S = \left\{ x_i, y_i \right\}_{i=1}^{B}$, we select the negative pairs as samples with different labels ($N(i) \equiv \left\{x_j \in S, y_j\not=y_i \right\}$).
\paragraph{Hard selection.} Given a batch $S = \left\{ x_i, y_i; s_i, d_i \right\}_{i=1}^{B}$, where $s_i$ and $d_i$ denote the script and domain of $x_i$, negative pairs are chosen as follows: for each sample $x_i$ in the batch, we select negative pair samples with a different label $y$, the same script $s$, and the same domain $d$. If fewer than $K$ such negatives exist, we relax the conditions in the following order: (2) different label $y$ and same script $s$, (3) different label $y$ and same domain $d$, (4) different label $y$ only (\ie, {equivalent to Soft Selection}). The algorithm of hard selection method is given in Appendix~\ref{appendix:hard_selection_algorithm}.

\subsection{Memory Bank}
\label{sec:memory_bank}

The performance of Supervised Contrastive Learning (SCL) is highly sensitive to batch size, as both positive and negative pairs are sampled within each mini-batch. So, large and diverse batches are required to increase the effectiveness of SCL where the ideal case is having all labels in each batch.

In most studies employing SCL, the number of classes is relatively small (typically fewer than 10), allowing for batch sizes in the range of 64 to 512. However, in our case, the number of classes is substantially larger (\ie, 2,099), surpassing the feasible batch sizes that can be accommodated by contemporary GPUs. To mitigate this limitation, and following the work by \citet{tan-etal-2022-domain}, we integrate a \textit{memory bank}, that stores the last $M$ data samples in memory in addition to the current batch ($B$), allowing each data sample in the batch to draw the negative and positive samples from $B+M$ samples. This increases the number of possible negative and positive samples making the contrastive loss more representative. If the number of stored examples exceeds $M$, only the most recent $M$ examples are retained, while older examples are discarded. Then, $\mathcal{L}_{SCL}$ is computed using $M+B$ examples.

%% file: sections/experimental_setup.tex
\section{Experimental Setup}
\label{sec:experiments}

We adopt the same text representation approach as FastText, utilizing character n-grams and word embeddings while following the same hyperparameter settings established by \citep{nllbteam2022languageleftbehindscaling}.
\subsection{Training}
\label{sec:train_details}

\paragraph{Dataset.} We train our model using the \glotlidc{} dataset \citep{kargaran-etal-2023-glotlid} covering 2,099 languages. As no official splits were released by the authors, we partition the dataset into training (85\%) and testing (15\%) subsets. To address the imbalance between high- and low-resource languages, we down-sample high-resource languages in the training set to a maximum of 100k sentences per language. This down-sampling reduces the total number of training sentences from 306M to 62M, impacting 301 high-resource languages. We denote this down-sampled version of the training dataset as \glotlidctrain. We report the detailed data distribution in Appendix~\ref{appendix:data-stats}.

\paragraph{Models.}

We train the following LID models using \glotlidctrain{}: \\
(a) \textbf{\CE} - The model trained only with the cross-entropy objective, as defined in Equation 
\eqref{eq:cross_entropy}.\\
(b) \textbf{\SCL} - The model trained with the combined objective in Equation \eqref{eq:both}, without employing a memory bank.\\
(c) \textbf{\MS} - The model trained with the combined objective in Equation \eqref{eq:both}, employing a memory bank with soft negative pair selection, as explained in \S\ref{sec:negative_pair_selection}.\\
(d) \textbf{\MH} - The model trained with the combined objective in Equation \eqref{eq:both}, employing a memory bank with hard negative pair selection, as explained in \S\ref{sec:negative_pair_selection}.

\noindent We set the memory bank size to $M = 2048$ for both soft selection ($M_S$) and hard selection ($M_H$), with a minimum number of negatives $K = 1024$ for hard selection. In our experiments, we use a batch size of $B = 128$. The details of our hyperparameter tuning process are provided in Appendix~\ref{appendix:hyper-parameters}.

To further improve inference performance, we employ an ensemble approach that integrates the predictions of \CE{} and \MS{} by selecting the maximum their probabilities, denoted as \textbf{\MSensemble}. We report the ensembling method results for \MS{} only as we did not observe significant differences when using \MH{}.

\paragraph{Baselines.} In our experiments, we implemented \CE{} as a replication of \glotlidm{} \cite{kargaran-etal-2023-glotlid}, as official data splits for \glotlidm{} were not openly released (hindering a non-contaminated evaluation on \glotlidctest).
However, to enable a full comparison with prior methods, we also report the performance of the open-weight \glotlidm{}, the previous SOTA model on \udhr{}, and \textbf{\MGensemble{}}, our ensemble of \glotlidm{} and \MS{} by taking the sum of their predicted distributions. Technical details can be found in Appendix~\ref{appendix:sota_comparison_details}.

We also compare \MS{} against two other publicly available LID models: AfroLID \citep{adebara-etal-2022-afrolid} and NLLB-LID \citep{nllbteam2022languageleftbehindscaling}.

\label{sec:eval_metrics}

\subsection{Evaluation}
We evaluate our models on the \glotlidctest{} \citep{kargaran-etal-2023-glotlid}, \flores{} \cite{nllbteam2022languageleftbehindscaling}, and
\udhr{} (Universal Declaration of Human Rights) datasets.

\paragraph{\glotlidctest.} This dataset is comprised of 15\% of the examples in \glotlidc{} (described above, \S\ref{sec:train_details}), which contains 2,099 languages.

\paragraph{\flores.} This dataset comprises 842 articles sourced from English-language Wikimedia projects. Each sentence in the dataset has been translated into 204 unique language-script combinations, covering 196 distinct languages. These translations have been manually verified by human annotators to ensure quality and accuracy.
Since the FLORES-200 development set is incorporated into \glotlidc, and following prior work, we adopt the dev-test subset (containing 1,012 sentences) as our evaluation set for \texttt{FLORES-200}. Our analysis focuses on 199 languages that overlap between FLORES-200 and \glotlidc.

\input{tables/overal_performance}

\paragraph{\udhr.} The original \udhr{} dataset contains over 500 translations of the Universal Declaration of Human Rights. \citet{kargaran-etal-2023-glotlid} curated and refined this dataset, reducing it to 369 languages. During our experiments, we identified discrepancies in seven languages (dip\_Latn, pcd\_Latn, aii\_Syrc, guu\_Latn, mto\_Latn, chj\_Latn, auc\_Latn) due to character misalignments between \glotlidc{} and \udhr, and two languages (taj\_Deva, tzm\_Tfng) that had only one example each. We exclude these nine languages, resulting in a final \udhr{} dataset comprising 360 languages.
\udhr{} serves as an out-of-domain evaluation dataset and is not included in \glotlidctrain.

\paragraph{\finewebtwo .} This dataset is a multilingual pretraining corpus encompassing $\sim$1,800 languages \citep{penedo2024fineweb2}. The LID process in the \finewebtwo{} pipeline is conducted using \glotlidm{} \cite{kargaran-etal-2023-glotlid}, a model capable of recognizing 2102 languages. 
We use this dataset to evaluate our model in large-scale real-world scenarios, which provides insights into the practical significance of our approach for LID for web crawls.

\paragraph{Metrics.} Following previous work \cite{nllbteam2022languageleftbehindscaling, kargaran-etal-2023-glotlid}, we report the F1 score and the false positive rate (FPR) as evaluation metrics. The F1 score combines precision and recall, both critical metrics: precision reflects the accuracy of classification decisions, while recall ensures minimal data loss. FPR, defined as $\text{FPR} = \frac{\text{FP}}{\text{FP}+\text{TN}}$, where FP represents false positives and TN represents true negatives, evaluates the impact of false positives. This is particularly important in our scenario, where the negative class is significantly larger, amplifying the effects of even a low false positive rate.

\input{tables/language_family_performance}
\input{tables/domain_performance}

%% file: tables/overal_performance.tex
\begin{table*}
  \centering
  \resizebox{0.9\textwidth}{!}{
  \begin{tabular}{lcc cc cc}
    \toprule
    \textbf{Method} & \multicolumn{2}{c}{\textbf{GlotLID-C test}} & \multicolumn{2}{c}{\textbf{UDHR}} & \multicolumn{2}{c}{\textbf{FLORES-200}} \\
     & \multicolumn{2}{c}{\textbf{(2099 languages)}} & \multicolumn{2}{c}{\textbf{(360 languages)}} & \multicolumn{2}{c}{\textbf{(199 languages)}} \\
    \cmidrule(lr){2-7}
         & \textbf{F1↑} & \textbf{FPR↓} & \textbf{F1↑} & \textbf{FPR↓} & \textbf{F1↑} & \textbf{FPR↓} \\

    \midrule
    \CE & .9858\textsubscript{6e-4} & .0000068\textsubscript{1.2e-7} & .8925\textsubscript{19e-4} & .0002222\textsubscript{34e-7} & .9650\textsubscript{3e-4} & {.0001142\textsubscript{9.9e-7}} \\
    \SCL & .9833\textsubscript{1e-4} & .0000075\textsubscript{0.2e-7} & .8938\textsubscript{19e-4} & \underline{.0002095\textsubscript{18e-7}} & .9617\textsubscript{3e-4} & .0001182\textsubscript{25e-7} \\
    \MS & .9860\textsubscript{2e-4} & .0000067\textsubscript{0.3e-7} & \underline{.9012\textsubscript{21e-4}} & {.0002170\textsubscript{28e-7}} & {.9652\textsubscript{3e-4}} & .0001177\textsubscript{20e-7} \\
    \MH & \underline{.9862\textsubscript{1e-4}} & \underline{.0000067\textsubscript{0.2e-7}} & .8961\textsubscript{20e-4} & .0002211\textsubscript{34e-7} & .9651\textsubscript{3e-4} & .0001174\textsubscript{14e-7} \\
    \MSensemble & \textbf{.9868\textsubscript{1e-4}} & \textbf{.0000065\textsubscript{0.4e-7}} & {.8998\textsubscript{15e-4}} & .0002180\textsubscript{33e-7} & {.9655\textsubscript{2e-4}} & {.0001171\textsubscript{19e-7}} \\
    \midrule    
        \glotlidm & - & - & .8919 & {.0002139} & \underline{.9696} & \underline{.0001071} \\
        \MGensemble & - & - & \textbf{.9060\textsubscript{15e-4}} & \textbf{.0002082\textsubscript{33e-7}} & \textbf{.9716\textsubscript{2e-4}} & \textbf{.0001025\textsubscript{19e-7}} \\

    \bottomrule
  \end{tabular}
  }
  \caption{Performance of LID models on \glotlidctest, \udhr{}, and \flores{} with 2,099, 360, and 199 languages respectively, when trained using \glotlidctrain{} dataset.
  Results are averaged over five runs and each subscript indicates the standard deviation of five runs. The best performance among the 5 methods is in \textbf{bold}, while the second best performance is \underline{underlined}. No results are reported for \glotlidm{} and \MGensemble{} on \glotlidctest{} as they were trained on splits that include parts of \glotlidctest{} in the training data.}
  \label{tab:lid-results}
\end{table*}

%% file: tables/language_family_performance.tex
\begin{table*}[t]
    \centering
    \small
    \begin{tabular}{ll!{\vrule}ccccc}
        \toprule
        \textbf{Resource Level} & \textbf{$|L|$} & \textbf{\CE} & \textbf{$\Delta$\SCL} & \textbf{$\Delta$\MS} & \textbf{$\Delta$\MSensemble} \\
        \midrule
        High-resource & 330 & 0.9065 & 0 & 0.0066 & 0.0066 \\
        Low-resource & 30 & 0.7380 & 0.0168 & 0.0323 & 0.0154 \\
        \bottomrule
    \end{tabular}
    \caption{F1 score change on \udhr{} of the LID models with SCL stratified by language resource level. \CE{} is used as a baseline, and subtracted from each model's score to compute the $\Delta$ score, \eg{}, $\Delta$\SCL = \SCL-\CE. $|L|$ refers to the number of languages.}% Scores of the \CE{} is also given in the last column as a reference.}
    \label{tab:analysis_by_lang_family}
\end{table*}

%% file: tables/domain_performance.tex
\begin{table*}[t]
    \centering
    \small
    \begin{tabular}{ll!{\vrule}ccccc}
        \toprule
        \textbf{Train-set Domains} & \textbf{$|L|$} & \textbf{\CE} & \textbf{$\Delta$\SCL} & \textbf{$\Delta$\MS} & \textbf{$\Delta$\MSensemble} \\
        \midrule
        \textbf{Random}  &  7    & 0.8157 & 0.0109 & 0.0541 & 0.0329          \\
        \textbf{Bible}   &  95   & 0.7821 & -0.0077 & 0.0190 & 0.0149         \\
        \bottomrule
    \end{tabular}
    \caption{F1 score change of the LID models' with SCL by GlotLID-C train domains. \CE{} is used as a baseline, and subtracted from each model's score to compute the absolute $\Delta$ score, \eg{} $\Delta$\SCL = \SCL-\CE. $|L|$ refers to the number of languages.} 
    \label{tab:analysis_by_domain}
\end{table*}

%% file: sections/analysis.tex
\section{Experimental Analysis}
\label{sec:analysis}

Table~\ref{tab:lid-results} presents the overall results for \glotlidctest, \udhr, and \flores. We observe no significant improvement when directly applying the SCL objective (\SCL) compared to the cross-entropy baseline (\CE). However, when incorporating a memory bank to increase the number of contrastive negatives beyond the number of classes, we achieve performance gains across all three evaluation datasets. Notably, this improvement is most significant for \udhr{} (out-of-domain), indicating a better generalization of the contrastive learning approach.

For in-domain evaluation (\glotlidctest{} and \flores), the highest F1 score among our methods is an ensemble approach that combines \CE{} and \MS{} with a memory bank of $\text{M}_\text{H}$=2048 and K=1024, suggesting that LID\textsubscript{CE} already performs well for certain languages (as we will show in \S\ref{sec:analysis_comparison}), while \MSensemble{} balances the performance across all labels.

When comparing to prior work, we observe that \glotlidm{} tends to outperform our methods on \flores, but underperforms them on \udhr, which is slightly more out of distribution. The ensembled method (\MGensemble{}), however, achieves the best score across \udhr{} and \flores{}, indicating the methods are complementary and could be combined for optimal accuracy. No comparison is made on \glotlidctest{} as \glotlidm{} was trained on different subsets of \glotlidc{}, meaning the evaluation would be contaminated (same for \MGensemble{}).

Table~\ref{tab:nllb_lid_afrolid} presents a performance comparison with AfroLID and NLLB-LID, evaluated on the subset of languages covered by all models. Results indicate that \MS{} consistently surpasses both models across all three benchmark datasets.
\input{tables/baselines}

\subsection{Performance on \udhr{} Dataset}
\label{sec:analysis_comparison}
We present a detailed analysis of the LID models on the \udhr{} dataset to systematically examine the specific aspects in which the SCL objective enhances performance and the conditions under which it may lead to performance degradation. 

\paragraph{Resource Analysis.}
\label{sec:by_language_resource}

As the performance of a language model is strongly influenced by the amount of training data available, we categorize languages into high- and low-resource groups based on the number of training examples available in \glotlidctrain. Languages with fewer than 10k examples are designated as \textbf{low-resource}, while those with 10k or more examples are categorized as \textbf{high-resource}.

For each group of languages, we compute the average F1 score difference between SCL-based models \textbf{(}\SCL, \MS, \MSensemble\textbf{)} and \CE. The results are presented in Table~\ref{tab:analysis_by_lang_family}.
Our analysis indicates that the model \MS{} yields the most notable improvements for low-resource languages, with F1 score gain of 3.23 percentage points, validating the importance of memory bank (\S\ref{sec:memory_bank}). In contrast, the performance gain for high-resource languages is more modest, at 0.66 percentage points.
We further extend this analysis in Appendix~\ref{appendix:resource_level_categorization}, where we present detailed results based on the low- and high-resource categorization defined by \citet{joshi-etal-2020-state}.

\paragraph{Domain Analysis.}
\label{sec:by_domain}

The \glotlidc{} dataset comprises five distinct textual domains: \textbf{Bible} Translations, \textbf{Literature} (\ie, stories \& books), \textbf{Politics} (\ie, government documents), \textbf{Multi-domain} (News and Wikipedia articles), and \textbf{Random} (machine translation datasets, chat logs, and web-scraped data). To evaluate the impact of SCL across different training data domains, we analyze languages trained exclusively on a single domain. While the distribution of \glotlidctrain{} is shown in Appendix~\ref{appendix:data-stats}, among the languages in \udhr{}, 7 languages have been exclusively trained on \textbf{Random}, while 95 on \textbf{Bible}.
Therefore, we compute the F1 score difference between SCL-based models \textbf{(}\SCL, \MS, \MSensemble\textbf{)} and \CE{} for random and bible domains and present the results in Table~\ref{tab:analysis_by_domain}.
Our results indicate that the inclusion of a memory bank leads to performance improvements across both domains. Notably, the \textbf{Random} datasets exhibit a more significant gain, with improvements of up to 5.41 percentage points.
We hypothesize that this is due to the broader range of subdomains covered by \textbf{Random}, which facilitate learning more generalized language representations. In contrast, the \textbf{Bible} domain is limited to religious texts, making it less conducive to generalization.

\paragraph{Script Analysis.}
\label{sec:by_script}
\input{tables/script_performance}

We conduct a similar analysis on the scripts of the languages and present the results in Table~\ref{tab:analysis_by_script}. While the \udhr{} dataset comprises 31 scripts, we report only those for which at least one of the metrics $\Delta$\SCL, $\Delta$\MS, $\Delta$\MSensemble{} is nonzero. As observed, \MS{} exhibits an improvement of up to  10.9 percentage points (on \textit{Java}), while some high-resource scripts show non-significant improvement (\eg, \textit{Hani}, \textit{Cyrillic}). We attribute this primarily to the resource level of the languages associated with each script, which aligns with our findings from \S\ref{sec:by_language_resource}, particularly regarding the behavior of low-resource languages. A breakdown analysis is given in the Appendix~\ref{appendix:analysis_by_script_resource}.

\subsection{Out-of-Domain Generalization}
\label{appendix:analysis_on_bible}
As noted \glotlidctrain{} contains the following 5 domains: Bible, Literature, Politics, Multi-domain, Random.
To further evaluate the out-of-domain generalization capabilities of our approach, we conducted the following experiment: we retained only the Bible domain from \glotlidctrain{} and excluded the others. Using this single-domain dataset, we trained two models --- one with SCL (\MSB) and one with CE (\CEB).
We then evaluated both models on all five domains of \glotlidc{}, as well as on Flores-200 and UDHR. The resulting macro F1-scores are presented in Table~\ref{tab:bible_experiment}.

The results demonstrate that even when training is limited to a single domain (\ie, Bible), our method improves generalization to previously unseen domains. Although the gains are modest under single-domain training, the advantage becomes substantially more pronounced when the model is trained on multiple domains.

\subsection{\finewebtwo{} Analysis}
To assess the scalability and real-world applicability of our approach as an LID tool for multilingual web crawls, we evaluate its performance on the \finewebtwo{} corpus.
\paragraph{Prediction Agreement.}
We compare \MS{} with \glotlidm{}, a state-of-the-art LID method, using a subset\footnote{10 -- 10k documents per language.} of the \finewebtwo{} corpus that has been labeled by \glotlidm{}. To quantify their similarity, we compute the agreement score between their predictions and report the results for high- and low-resource languages in Table~\ref{tab:agreement_fineweb2}. The agreement score is defined as follows:
$$\text{Agreement} = \frac{|\text{Pred. \textsubscript{\MS}} \cap \text{Pred. \textsubscript{\glotlidm{}}}|}{\text{\# of documents}}$$

\normalsize

\input{figures/analysis_domain}

\input{tables/fineweb2_agreement}

In \S\ref{sec:analysis_comparison}, we demonstrated that the incorporation of SCL was particularly beneficial for low-resource languages. To illustrate its effect, Table~\ref{tab:agreement_fineweb2} presents the agreement scores for different language groups in \finewebtwo{}. As expected, the agreement is high for high-resource languages, but much lower for low-resource languages, with a micro agreement rate of $\sim$58.61\%. Since the dataset lacks ground-truth labels, it is not possible to directly determine which model is correct in cases of disagreement. However, as shown earlier in \S\ref{sec:analysis_comparison}, incorporation of SCL outperforms the CE-based method, including \glotlidm{}, on low-resource languages, suggesting that a low agreement score in these languages likely reflects \MS{} making the correct predictions where the \glotlidm{} fails. Importantly, even small differences are practically meaningful: a 1\% improvement in low-resource language performance corresponds to $\sim$28,000 documents. This highlights the real-world impact of even minor improvements in low-resource LID models.

\section{Under-performing languages}
\label{sec:under_performance_analysis}

Using the \MS{} model, we conduct a detailed analysis of the characteristics of languages exhibiting low performance on the \udhr{} dataset (\ie, languages for which models display limited domain generalization capabilities).

\paragraph{Language Analysis by Domain.}

We categorize languages based on domain availability and the number of sentences available in \texttt{GlotLID-C-train-down}. Figure~\ref{fig:analysis_domains} presents the corresponding F1 score performance on the \udhr{} dataset. We observe that models tend to exhibit improved generalization (\ie, higher F1 scores on \udhr{}) in languages for which they had more diverse training domains. Models tend to generalize worse on languages for which training data was only available in one domain. Interestingly, while a majority of single-domain languages only had training data for the \textbf{Bible} domain, the model still achieves relatively high F1 scores for these languages, suggesting that training only on \textbf{Bible} data is not a sole factor contributing to low performance on \udhr{}. Finally, we observe that the size of the training data does not appear to be a sole determinant of low performance on \udhr{} either.

\begin{figure}
    \centering
    \includegraphics[width=1\linewidth]{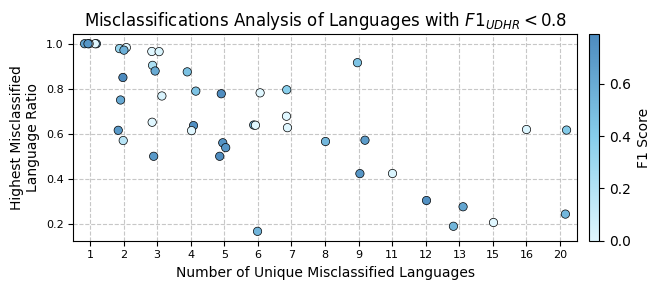}
    \caption{
    \udhr{} misclassification pairs (languages with F1 $<$0.8) using \MS{}.
    Most misclassification for a language happens between less than 5 languages.
    }
    \label{fig:analysis_missclass}
\end{figure}

\paragraph{Language Misclassifications.}
\label{sec:misclassifications}

To qualitatively assess these generalization issues, we now analyze patterns observed in misclassified language pairs. Among the 3448 misclassifications in the \udhr{} dataset, 99.6\% occur within the same script, while only 0.4\% involve different scripts.
Most misclassifications also involve a target language being incorrectly identified as another high-resource language (see Appendix Table~\ref{tab:analysis_language_family_direction}).

To better understand the diversity of languages involved in misclassifications, we examine the number of distinct languages a given language is misclassified as and whether there are consistent confusion patterns with specific languages. For instance, the language ``qug\_Latn'' (Chimborazo Highland Quichua), which belongs to the Quechuan language family and is spoken in Ecuador, is misclassified as 15 different languages. Notably, 60\% of these cases involve ``qup\_Latn'' (Southern Pastaza Quechua), another Quechuan language spoken in Ecuador, indicating a strong linguistic similarity that leads to frequent misclassification.

For languages in the \udhr{} dataset with an F1 score below 0.8, we visualize the misclassification trends in Figure~\ref{fig:analysis_missclass}. Specifically, we plot the ratio of the most frequently misclassified language to the total number of misclassifications for each language against the number of unique languages it is misclassified as.

Our analysis reveals that most languages are predominantly misclassified as only a small subset of the 2099 total languages (\ie, \textit{number of unique misclassified languages} < 10). This suggests that these languages share strong linguistic similarities with a limited set of other languages. Examples include Nyemba and Mbunda, both Bantu languages, and Northeastern and Southwestern Dinka, which exhibit significant linguistic overlap. Even languages with a high overall misclassification rate (\textit{number of unique misclassified languages} > 10) tend to have one or a few dominant misclassification sources (\ie, \textit{highest misclassified language ratio}), as seen in the case of Chimborazo Highland Quichua and Southern Pastaza Quechua.
More examples of such pairs are provided in Table~\ref{tab:lang_misclassification}.

%% file: tables/baselines.tex
\begin{table*}
  \centering
  \resizebox{0.6\textwidth}{!}{
  \begin{tabular}{lcc cc cc}
    \toprule
    \textbf{Method} & \multicolumn{2}{c}{\textbf{GlotLID-C test}} & \multicolumn{2}{c}{\textbf{UDHR}} & \multicolumn{2}{c}{\textbf{FLORES-200}} \\
    \cmidrule(lr){2-7}
         & \textbf{F1↑} & \textbf{FPR↓} & \textbf{F1↑} & \textbf{FPR↓} & \textbf{F1↑} & \textbf{FPR↓} \\
    
    \midrule
    \MS{}\textsuperscript{a} & \textbf{.9860} & \textbf{.0000058} & \textbf{.9012} & \textbf{.0002170} & \textbf{.9652} & \textbf{.0001177} \\
    AfroLID\textsuperscript{a} & .8609 & .0002883 & .7605 & .0012766 & .7618 & .0011196 \\
    \midrule
    \MS{}\textsuperscript{b} & \textbf{.9841} & \textbf{.0000058} & \textbf{.9639} & \textbf{.0002170} & \textbf{.9861} & \textbf{.0001177} \\
    NLLB-LID\textsuperscript{b} & .9335 & .0002380 & .9405 & .0003649 & .9614 & .0002708 \\

    \bottomrule
  \end{tabular}
  }
  \caption{Comparison of \MS{} with AfroLID and NLLB-LID systems. \textsuperscript{a}Evaluated on 398/82/44 languages for GlotLID-C test/UDHR/FLORES-200 respectively. \textsuperscript{b}Evaluated on 207/162/182 languages for GlotLID-C test/UDHR/FLORES-200 respectively.}
  \label{tab:nllb_lid_afrolid}
\end{table*}

%% file: tables/script_performance.tex
\begin{table*}[th]
    \centering
    \resizebox{0.6\textwidth}{!}{
    \begin{tabular}{lccccc}
        \toprule
        \textbf{Script}  & \textbf{\CE} & \textbf{$\Delta$\SCL} & \textbf{$\Delta$\MS} & \textbf{$\Delta$\MSensemble} \\
        \midrule
        \textbf{Hani}         & 0.5678 & -0.0265 & 0.0020 & 0.0011          \\
        \textbf{Tifinagh}     & 0.7158 & 0.0134  & 0.0134 & 0.0134          \\
        \textbf{Arabic}       & 0.8686 & 0.0277  & 0.0678 & 0.0325          \\
        \textbf{Latin}        & 0.8803 & 0.0012  & 0.0092 & 0.0079          \\
        \textbf{Java}         & 0.8829 & 0.0923  & 0.1090 & 0.1007          \\
        \textbf{Tibetan}      & 0.9411 & 0.0253  & 0.0084 & 0.0084          \\
        \textbf{Cyrillic}     & 0.9731 & 0.0053  & 0.0011 & 0.0030          \\
        \bottomrule
    \end{tabular}
    }
    \caption{F1 score change of the LID models' with SCL by language script. \CE{} is used as a baseline, and subtracted from each model's score to compute the absolute $\Delta$ score, \eg{} $\Delta$\SCL = \SCL-\CE.}
    \label{tab:analysis_by_script}
\end{table*}

%% file: figures/analysis_domain.tex
\begin{figure*}[t]
    \centering
    \includegraphics[width=0.8\linewidth]{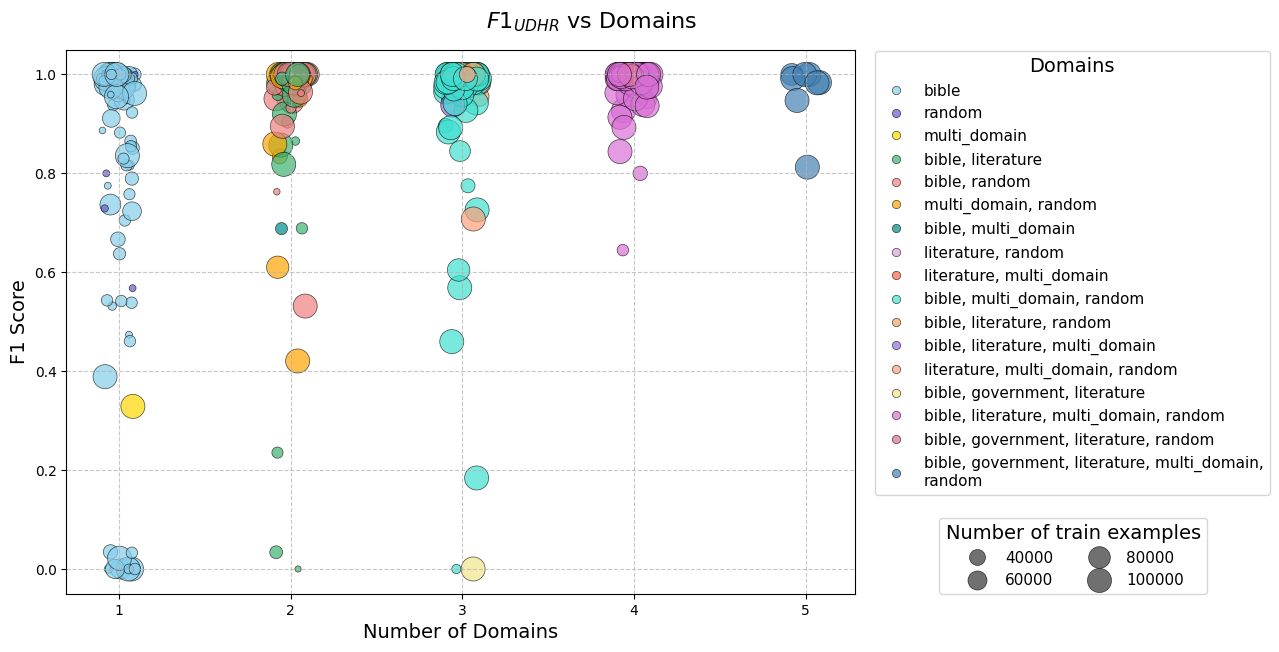}
    \caption{Performance of \MS{} on the UDHR dataset based on training data domains. The more domains included in the training data, the higher performance the model shows for a given language. The size of the training data is not always positively correlated with model's performance, particularly in cases where languages are represented in limited domains.
    % by the number of domains and the number of examples in \texttt{GlotLID-C train}.
    }
    \label{fig:analysis_domains}
\end{figure*}

%% file: tables/fineweb2_agreement.tex
\begin{table}[t]
    \centering
    \resizebox{\linewidth}{!}{

    \begin{tabular}{lll!{\vrule}cc}
        \toprule
        \textbf{Resource} & \multirow{2}{*}{${ \vert L \vert}$}&  \multirow{2}{*}{${\vert D \vert}$} & \textbf{Macro} & \textbf{Micro} \\
        \textbf{Level} &  & & \textbf{Agreement} & \textbf{Agreement} \\
        \midrule
        High & 1396 & 4.5B & 0.9004 & 0.9064 \\
        Low & 273 & 2.9M & 0.6708 & 0.5861 \\
        \bottomrule
    \end{tabular}
    }
    \caption{Agreement between \MS{} and \glotlidm{} on \finewebtwo{} filtered dataset. $|L|$ refers to the number of languages, while $|D|$ refers to the total number of documents in each group. The agreement scores have been calculated using 10--10K documents per language.}
    \label{tab:agreement_fineweb2}
\end{table}

%% file: sections/conclusion.tex
\section{Conclusion}

In this work, we propose a novel approach that leverages supervised contrastive learning (SCL) for language identification (LID). The SCL objective explicitly encourages textual representations of the same language to form compact clusters while pushing apart representations of different languages in the embedding space. We evaluate our approach on three benchmark datasets: GlotLID, FLORES, and UDHR, demonstrating that incorporating SCL leads to considerable improvements over conventional cross-entropy (CE)-based LID methods. Specifically, our best SCL-based models achieve a 3.2 percentage points improvement over CE-based models for low-resource languages and a 6.8 percentage points improvement for languages with data from diverse domains, highlighting improved domain generalization, and underscoring the effectiveness of SCL in enhancing LID performance, particularly in challenging low-resource and cross-domain scenarios. Moreover, we also demonstrate that SCL and CE-based methods are complementary to each other and can be used together during inference, as in the case of \MGensemble{}.

%% file: sections/limitations.tex
\section*{Limitations}
% one limitation of SCL is being slower than a linear classifier and as we use LID in lareg-scale (web-scale), being fast matters a lot
% 
Contrastive learning has been shown to enhance model performance and generalization by capturing a broader range of factors of variation compared to traditional classification networks. However, its effectiveness relies on access to high-quality data from diverse domains, which poses a challenge in the development of Language Identification (LID) systems, where such data is often scarce.

% Additionally, contrastive learning is highly sensitive to the selection of positive and negative examples, making the training process more complex. As a result, extensive ablation studies and careful hyperparameter tuning are necessary to optimize its performance and ensure its practical applicability.

While our training and in-domain test datasets encompass 2,099 languages, our out-of-domain evaluation dataset (\udhr) is limited to only 360 languages. This restriction constrains our ability to conduct a comprehensive analysis across the remaining languages. Additionally, the lack of evaluation dataset in other domains limits our ability to thoroughly assess method's generalization capacity.

%% file: sections/appendix_hyperparameters.tex
\section{Data Statistics}
\label{appendix:data-stats}

We report the \glotlidctrain{} distribution by domains (Table~\ref{tab:script_stats}), script (Table~\ref{tab:train_domains_stats}) and resource level (Table~\ref{tab:resource_level_stats}). As can be seen from Table~\ref{tab:train_domains_stats}, most languages contains either exclusively or inclusively Bible data. 

\begin{table}[ht]
    \centering
    \small
    \begin{tabular}{lcc}
        \toprule
        \textbf{Script} & \textbf{\#Languages} & \textbf{Avg. \#Data Samples} \\
         &  & \textbf{per language} \\
        \midrule
        Java  & 2    & 50,118 \\
        Tibt  & 4    & 42,773 \\
        Hani  & 5    & 83,411 \\
        Tfng  & 5    & 22,801 \\
        Arab  & 39   & 46,203 \\
        Cyrl  & 71   & 42,553 \\
        Latn  & 1702 & 30,706 \\
        \bottomrule
    \end{tabular}
    \caption{\glotlidctrain{} distribution by scripts. Only the scripts used in \S\ref{sec:by_script} are shown.}
    \label{tab:script_stats}
\end{table}

\begin{table}
    \centering
    \scriptsize
    \begin{tabular}{lcc}
        \toprule
        \textbf{Domains} & \textbf{\#Languages} & \textbf{Avg. \#Data Samples} \\
         &  & \textbf{per language} \\
        \midrule
        \textbf{Bible}, Government, Literature & 1 & 100,000 \\
        \textbf{Bible}, Government, \\ Literature, Random & 1 & 100,000 \\
        Literature, Random & 2 & 50,138 \\
        Literature, Multi-Domain & 2 & 11,742 \\
        \textbf{Bible}, Literature, Multi-Domain & 3 & 59,662 \\
        Literature & 4 & 525 \\
        \textbf{Bible}, Government, Literature, \\Multi-Domain, Random & 8 & 97,635 \\
        Literature, Multi-Domain, Random & 8 & 44,277 \\
        \textbf{Bible}, Literature, Random & 16 & 57,210 \\
        \textbf{Bible}, Multi-Domain & 21 & 25,936 \\
        Multi-Domain & 42 & 13,764 \\
        Multi-Domain, Random & 53 & 32,901 \\
        Random & 61 & 7,061 \\
        \textbf{Bible}, Literature, Multi-Domain, \\Random & 89 & 93,826 \\
        \textbf{Bible}, Random & 95 & 42,333 \\
        \textbf{Bible}, Multi-Domain, Random & 121 & 76,753 \\
        \textbf{Bible}, Literature & 157 & 35,128 \\
        UND & 161 & 100,000 \\
        \textbf{Bible} & 1254 & 22,874 \\ [0.2em]
        \hline \\ [-0.7em]
        Overall & 2099 & 51,145 \\
        \bottomrule
    \end{tabular}
    \caption{\glotlidctrain{} distribution by domains combination. Those with \textbf{Bible} data included are highlighted.}
    \label{tab:train_domains_stats}
\end{table}

\begin{table}[ht]
    \centering
    \small
    \begin{tabular}{lcc}
        \toprule
        \textbf{Resource Level} & \textbf{\#Languages} & \textbf{Avg. \#Data Samples} \\
         &  & \textbf{per language} \\
        \midrule
        Low  & 398  & 3,959 \\
        High & 1540 & 39,049 \\
        \bottomrule
    \end{tabular}
    \caption{\glotlidctrain{} distribution by resource level. Those with UND labels have been excluded.}
    \label{tab:resource_level_stats}
\end{table}

\section{Training Details}

\subsection{Comparison details to \glotlidm{}}
\label{appendix:sota_comparison_details}
The comparison of technical details of \MS{}, \CE{} and \glotlidm{} can be found from Table~\ref{tab:lid_config_comparison}.
\begin{table*}
    \centering
    \small
    \begin{tabular}{lccc}
        \toprule
        \textbf{} & \textbf{\glotlidm{}} & \textbf{\CE{}} & \textbf{\MS{}} \\
        \midrule
        Train data       & \glotlidc (85\%)       & \glotlidc (85\%)  & \glotlidc (85\%)\\
               & (Upsampled)       & (Downsampled to 100K) & (Downsampled to 100K) \\
        Loss function    & CE                     & CE & CE+SCL                         \\
        Number of labels & 2102                              & 2099 & 2099                                 \\
        Train Module     & FastText                          & PyTorch & PyTorch                \\
        Optimizer        & SGD                               & ADAM & ADAM                                  \\
        \#Epochs      & 2                                 & 1 & 1                                     \\
        Learning Rate    & 0.8                               & 1e-3 & 1e-3                                  \\
        \bottomrule
    \end{tabular}
    \caption{Comparison of \MS{}, \CE{} and \glotlidm{}  training configurations.}
    \label{tab:lid_config_comparison}
\end{table*}

\subsection{Hyperparameters}
\label{appendix:hyper-parameters}

We adopt the same text representation approach as FastText, utilizing character n-grams and word embeddings . Specifically, we set the following parameters: \textit{minCount} = 1000, \textit{minCountLabel} = 0, \textit{wordNgrams} = 1, \textit{bucket} = $10^6$, \textit{minn} = 2, \textit{maxn} = 5, and \textit{dim} = 256.

We train our models for a single epoch following \citet{kargaran2024glotcc}. We, also, found that additional epochs yield no noticeable improvement due to the rapid convergence. We use a batch size of 128 with a gradient accumulation step of 1, which is the maximum supported by our GPU configuration. Model training was conducted on 8 NVIDIA A100 GPUs (80 GB each) with 512 GB RAM. Optimization was performed using the AdamW optimizer with the following hyper-parameters: $\beta_1=0.9$, $\beta_2=0.999$, $\epsilon=1e-06$, weight decay $\lambda=0$, and a linear learning rate schedule.

Hyper-parameter tuning was conducted on the GlotLID-C-test, FLORES-200, and UDHR datasets, involving all 369 languages for UDHR. Macro F1 score served as the primary evaluation metric. Below, we summarize the tuning process and the selected hyper-parameters for down-sampling, contrastive temperature ($\tau$), learning rate, memory bank sizes ($M_S$, $M_H$), and the minimum number of negatives ($K$).

\begin{table}
    \centering
    \small
    \begin{tabular}{lcccc}
        \toprule
         & \multicolumn{3}{c}{{Down-sampling max cap}} & \\ 
        % \midrule
        \cmidrule(lr){2-5}
         & \textbf{10K} & \textbf{50K} & \textbf{100K} & \textbf{Full} \\
        \midrule
        GlotLID-C-test & 0.9776 & 0.9855 & 0.9868 & 0.9855 \\
        UDHR & 0.8592 & 0.8683 & 0.8688 & 0.8556 \\
        FLORES-200 & 0.9518 & 0.9615 & 0.9638 & 0.9621 \\
        \bottomrule
    \end{tabular}
    \caption{Macro F1 score of LID\textsubscript{CE} using {10K, 50K, 100K} to down-sample the high-resource languages, and using the full train dataset.}
    \label{tab:downsampling}
\end{table}

\begin{table}
    \centering
    \scriptsize
    \begin{tabular}{lccccc}
        \toprule
         & \multicolumn{5}{c}{\textbf{Contrastive Temperature } $\tau$} \\ 
        \cmidrule(lr){2-6}
         & \textbf{0.02} & \textbf{0.05} & \textbf{0.1} & \textbf{0.2} & \textbf{0.4} \\
        \midrule
        GlotLID-C-test & 0.9841 & 0.9864 & 0.9866 & 0.9851 & 0.9861 \\
        UDHR & 0.8634 & 0.8687 & 0.8687 & 0.8617 & 0.8635 \\
        FLORES-200 & 0.9602 & 0.9625 & 0.9628 & 0.9615 & 0.9624 \\
        \bottomrule
    \end{tabular}
    \caption{Macro F1 score of LID\textsubscript{SCL} using contrastive temperature, $\tau$, of {0.02, 0.05, 0.1, 0.2, 0.4}.}
    \label{tab:tau}
\end{table}

\begin{table}
    \centering
    \small
    \begin{tabular}{lccc}
        \toprule
         & \multicolumn{3}{c}{\textbf{Learning Rate}} \\ 
        \cmidrule(lr){2-4}
         & \textbf{1e-3} & \textbf{2e-3} & \textbf{4e-3} \\
        \midrule
        GlotLID-C-test & 0.9864 & 0.9858 & 0.9841 \\
        UDHR & 0.8687 & 0.8631 & 0.8552 \\
        FLORES-200 & 0.9625 & 0.9622 & 0.9611 \\
        \bottomrule
    \end{tabular}
    \caption{Macro F1 score of LID\textsubscript{SCL} trained with the learning rate values of {1e-3, 2e-3, 4e-3}.}
    \label{tab:learning_rate}
\end{table}

\begin{table}
    \centering
    \small
    \begin{tabular}{lccc}
        \toprule
         & \multicolumn{3}{c}{$\textbf{M}_\textbf{S}$} \\ 
        \cmidrule(lr){2-4}
         & \textbf{1024} & \textbf{2048} & \textbf{4096} \\
        \midrule
        GlotLID-C-test & 0.9884 & 0.9883 & 0.9883 \\
        UDHR & 0.8739 & 0.8742 & 0.8713 \\
        FLORES-200 & 0.9646 & 0.9649 & 0.9648 \\
        \bottomrule
    \end{tabular}
    \caption{Macro F1 score of \MS{} with the values of {1024, 2048, 4096} for memory bank size, $M_S$.}
    \label{tab:ms}
\end{table}

\begin{table}
    \centering
    \scriptsize
    \begin{tabular}{llcccc}
        \toprule
         & & \multicolumn{4}{c}{\textbf{$\textbf{M}_\textbf{H}$}} \\ 
        \cmidrule(lr){3-6}
         & \multicolumn{1}{c}{\textbf{K}} & \textbf{1024} & \textbf{2048} & \textbf{4096} & \textbf{8192} \\
        \midrule
        GlotLID-C-test & 512 & 0.9884 & 0.9882 & 0.9884 & 0.9882 \\
         & 1024 & - & 0.9883 & 0.9885 & 0.9885 \\
        UDHR & 512 & 0.8690 & 0.8731 & 0.8743 & 0.8700 \\
         & 1024 & - & 0.8739 & 0.8716 & 0.8710 \\
        FLORES-200 & 512 & 0.9649 & 0.9649 & 0.9651 & 0.9653 \\
         & 1024 & - & 0.9647 & 0.9648 & 0.9646 \\
        \bottomrule
    \end{tabular}
    \caption{Macro F1 score of \MH{} with the values \{1024, 2048, 4096, 8192\} and \{512, 1024\} for memory bank size ($M_H$) and minimum number of negatives ($K$) respectively.}
    \label{tab:mh_k}
\end{table}

\vspace{0.3cm}
\noindent
\textbf{GlotLID-C-train Down-sampling:} We evaluated down-sampling high-resource languages using values {10K, 50K, 100K}. The LID\textsubscript{CE} model performed best with a cap of 100K, as shown in Table~\ref{tab:downsampling}. Up-sampling, as proposed by \citet{nllbteam2022languageleftbehindscaling}, was also tested but yielded inferior results compared to down-sampling.

\noindent
\textbf{Effect of Contrastive Temperature ($\tau$):} We experimented with temperatures {0.02, 0.05, 0.1, 0.2, 0.4} (Table~\ref{tab:tau}) and selected 0.05 based on empirical results and alignment with findings from \citet{gao-etal-2021-simcse}.

\noindent
\textbf{SCL Learning Rate:} Initial experiments indicated that learning rates in the order of $10^{-3}$ achieved better convergence. We compared rates {1e-3, 2e-3, 4e-3} in Table~\ref{tab:learning_rate} and selected 1e-3 as the optimal value.

\noindent
\textbf{Memory Bank Size for Soft Selection ($M_S$):} We evaluated values {1024, 2048, 4096} for $M_S$ and selected 2048 as the final configuration (Table~\ref{tab:ms}).

\noindent
\textbf{Memory Bank Size ($M_H$) and Minimum Number of Negatives ($K$) for Hard Selection:} Since $M_H$ and $K$ are interdependent, we tested combinations as shown in Table~\ref{tab:mh_k}. The best performance was achieved with $M_H=2048$ and $K=1024$.

\input{tables/lang_family_direction}

\section{Hard Selection Algorithm}
\label{appendix:hard_selection_algorithm}

The negatives pairs in the Hard selection method, as explained in \S\ref{sec:negative_pair_selection}, are sampled according to the Algorithm~\ref{algo:hard_neg}.

\begin{algorithm}[t]
\caption{Hard Negative Sample Selection}
\begin{algorithmic}[1]
\small
\REQUIRE A minibatch $S = \left\{ x_i, y_i; s_i, d_i \right\}_{i=1}^{B}$, and a number of minimum negatives $K$
\FOR{$i < B$}
    \STATE $N(i) = \left\{z_j \in S, y_j \neq y_i, s_j = s_i, d_j = d_i \right\}$ \COMMENT{Condition 1}
    \IF{$|N(i)| < K$}
        \STATE $N(i) = \left\{z_j \in S, y_j \neq y_i, s_j = s_i\right\}$ \COMMENT{Condition 2}
    \ENDIF
    \IF{$|N(i)| < K$}
        \STATE $N(i) = \left\{z_j \in S, y_j \neq y_i, d_j = d_i\right\}$ \COMMENT{Condition 3}
    \ENDIF
    \IF{$|N(i)| < K$}
        \STATE $N(i) = \left\{z_j \in S, y_j \neq y_i\right\}$ \COMMENT{Condition 4}
    \ENDIF
\ENDFOR
\RETURN $N(i)$ \COMMENT{negative pairs for each sample}
\end{algorithmic}
\label{algo:hard_neg}
\end{algorithm}
% \vspace{-20pt} 
% \noindent $S$ is replaced by $S_C$ in the usage of memory bank, $Q$, as explained in \ref{sec:memory_bank}.

\input{tables/language_misclassification}

\section{Additional Results}

\begin{table*}[th]
    \centering
    \small
    \begin{tabular}{lccccc}
        \toprule
        \textbf{Dataset} & \textbf{Domain Shift} & \textbf{Resource Level} & $\lvert L_\text{test} \rvert$ & \textbf{\CEB} & \textbf{\MSB} \\
        \midrule
        Bible & ID        & High & 1400 & \textbf{0.9961} & \textbf{0.9961} \\
                    &      & Low  & 368  & 0.9883 & \textbf{0.9885} \\
        Random & OOD      & High & 202  & 0.5933 & \textbf{0.6075} \\
                    &      & Low  & 104  & 0.6429 & \textbf{0.6481} \\
        Multi-Domain  & OOD & High & 152  & 0.7996 & \textbf{0.8067} \\
                    &      & Low  & 90   & \textbf{0.7514} & 0.7444 \\
        Literature  & OOD  & High & 119  & 0.8521 & \textbf{0.8638} \\
                    &      & Low  & 51   & 0.7641 & \textbf{0.7890} \\
        Government  & OOD  & High & 5    & \textbf{0.9233} & 0.9137 \\
                    &      & Low  & 5    & 0.6263 & \textbf{0.6358} \\
        UDHR  & OOD        & High & 246  & 0.8854 & \textbf{0.8886} \\
                    &      & Low  & 83   & 0.7961 & \textbf{0.8039} \\
        FLORES-200  & OOD  & High & 101  & 0.8938 & \textbf{0.8969} \\
                    &      & Low  & 66   & 0.8367 & \textbf{0.8375} \\
        \bottomrule
    \end{tabular}
    \caption{Macro F1 scores of \MSB{} and \CEB{} across in-domain and out-of-domain datasets.}
    \label{tab:bible_experiment}
\end{table*}

\subsection{Significance testing using bootstrapping}
\label{appendix:significant_testing}

We additionally applied a t-test using bootstrapping as a significance testing method to more clearly demonstrate the effectiveness of SCL approach (Table~\ref{tab:bootstrapping}). Based on the bootstrapped results, p-value is less than 0.05, and the SCL method still outperforms both the \glotlidm{} and the cross-entropy baseline, \CE. 

The relatively lower scores for the UDHR dataset can be attributed to the imbalance in the number of sentences across languages. Some overrepresented languages exhibit lower performance, and their frequent selection during bootstrapping iterations negatively impacts the overall results.

\begin{table}
    \centering
    \scriptsize
    \begin{tabular}{lccc}
        \toprule
         & GlotLID-C Test & UDHR & FLORES-200 \\
        \midrule
        \CE{} & 0.9874 & 0.6080 & 0.8598 \\
        \MS{} & 0.9878 & 0.6176 & 0.8616 \\
        \glotlidm{} & 0.9877 & 0.5771 & 0.8614 \\
        \MGensemble{} & 0.9943 & 0.6201 & 0.8859 \\
        \bottomrule
    \end{tabular}
    \caption{Macro F1 score via bootstrapping with 1000 iterations (p<0.05)}
    \label{tab:bootstrapping}
\end{table}

\subsection{Script Analysis Breakdown}
\label{appendix:analysis_by_script_resource}

We showed in \S\ref{sec:by_script} that while low-resource scripts have shown significant improvement over the baseline using \MS{}, some high-resource scripts had more modest improvement. We show the break-down of the script analysis by the resource level in Table~\ref{tab:analysis_by_script_resource}.
We attribute the performance gains primarily to the resource level of the languages associated with each script.

While the number of low-resource languages is typically smaller, those within a given script tend to show greater improvements in performance compared to their high-resource counterparts.
Additionally, the number of languages per script varies significantly. For instance, the Latin script encompasses 282 languages in this table, whereas the substantial improvement seen in the Java script is based on a single language. This discrepancy in language count per script also contributes to the observed performance variance.

\begin{table*}[th]
    \centering
    % \small
    \resizebox{0.75\textwidth}{!}{
    \begin{tabular}{lcccccc}
        \toprule
        \textbf{Script} & \textbf{Resource Level} & \textbf{\#\udhr{} Languages} & \textbf{\CE} & \textbf{$\Delta$\SCL} & \textbf{$\Delta$\MS} & \textbf{$\Delta$\MSensemble} \\
        \midrule
        \textbf{Arabic}       & Low  & 5 & 0.5176 & 0.1275  & 0.2824 & 0.1419 \\
        \textbf{Arabic}       & High & 1 & 0.9388 & 0.0078  & 0.0249 & 0.0106 \\
        \textbf{Cyrillic}     & Low  & 5 & 0.9525 & 0.0072  & 0.0019 & 0.0038 \\
        \textbf{Cyrillic}     & High & 24 & 0.9877 & 0.0039  & 0.0005 & 0.0023 \\
        \textbf{Hani}         & High & 4 & 0.7570 & -0.0468 & -0.0088 & 0.0014 \\
        \textbf{Java}         & Low  & 1 & 0.8829 & 0.0923  & 0.1090 & 0.1007 \\
        \textbf{Latin}        & Low  & 20 & 0.8394 & 0.0039  & 0.0128 & 0.0108 \\
        \textbf{Latin}        & High & 262 & 0.9560 & -0.0037 & 0.0026 & 0.0025 \\
        \textbf{Tifinagh}     & Low  & 1 & 0.7158 & 0.0134  & 0.0134 & 0.0134 \\
        \textbf{Tibetan}      & High & 2 & 0.9411 & 0.0253  & 0.0084 & 0.0084 \\
        \bottomrule
    \end{tabular}
    }
    \caption{F1 score change of the LID models' with SCL by language script broken down by the resource level. \CE{} is used as a baseline, and subtracted from each model's score to compute the absolute $\Delta$ score, \eg{} $\Delta$\SCL = \SCL-\CE.}
    \label{tab:analysis_by_script_resource}
\end{table*}

% \subsection{Out-of-Domain Generalization}
% \label{appendix:analysis_on_bible}
% As noted \glotlidctrain{} contains the following 5 domains: Bible, Literature, Politics, Multi-domain, Random.
% To further evaluate the out-of-domain generalization capabilities of our approach, we conducted the following experiment: we retained only the Bible domain from \glotlidctrain{} and excluded the others. Using this single-domain dataset, we trained two models --- one with SCL (\MSB) and one with CE (\CEB).
% We then evaluated both models on all five domains of \glotlidc{}, as well as on Flores-200 and UDHR. The resulting macro F1-scores are presented in Table~\ref{tab:bible_experiment}.

% The results demonstrate that even when training is limited to a single domain (\ie, Bible), our method improves generalization to previously unseen domains. Although the gains are modest under single-domain training, the advantage becomes substantially more pronounced when the model is trained on multiple domains, as discussed in our analysis in Section~\ref{sec:analysis}.

\subsection{Resource level categorization}
\label{appendix:resource_level_categorization}

Languages can be categorized as low- or high-resource based on either the availability of large amounts of unlabeled data (\eg, pretraining data collected from the web) or the quantity of labeled data available for specific tasks.
In many cases, these two criteria coincide. However, in this study, our primary goal is to evaluate the effectiveness of our approach under conditions where labeled data for training an LID system is scarce. For this reason, we adopt the second criterion --- the amount of labeled training data --- as the basis for our categorization. Specifically, we classify languages with fewer than 10,000 labeled examples as low-resource, and those with 10,000 or more labeled examples as high-resource, as described in Section~\ref{sec:by_language_resource}. We find this definition particularly meaningful because the performance of language models is strongly influenced by the amount of labeled data available for training.
To provide a complementary perspective, we also analyze our results using the categorization proposed by \citet{joshi-etal-2020-state}. Table~\ref{tab:joshi_categorization} presents the performance of our method on the \udhr{} dataset according to this scheme. Labels 0 through 5 follow the definitions from \citet{joshi-etal-2020-state}, and we introduce an additional label, -1, for languages not included in their original categorization (low-resource and under-represented languages). As shown, most of the performance gains occur within these labeled groups, highlighting the robustness of our approach across different categorization schemes.

\begin{table}
    \centering
    \begin{tabular}{llcc}
        \toprule
        \textbf{Language Class} & \textbf{$|L|$} & \textbf{\CE{}} & \textbf{\MS{}} \\
        \midrule
        -1 & 135 & 0.8183 & 0.0204 \\
        0 & 93 & 0.9059 & 0.0004 \\
        1 & 76 & 0.9554 & 0.0036 \\
        2 & 13 & 0.9686 & -0.0024 \\
        3 & 20 & 0.9830 & 0.0067 \\
        4 & 17 & 0.9311 & -0.0020 \\
        5 & 6 & 0.9812 & 0.0008 \\
        \bottomrule
    \end{tabular}
    \caption{The performance of the models on \udhr{} dataset based on the categorization low- and high-resource languages of \citet{joshi-etal-2020-state}.}
    \label{tab:joshi_categorization}
\end{table}

\subsection{\textsc{UND} labels recovery}

\input{figures/fineweb2_recovery_analysis}
\citet{kargaran-etal-2023-glotlid} introduced the UND\_XXX labels due to the nature of GlotLID-M, which is based on the FastText model. This model relies on hashed n-grams rather than storing the n-grams themselves to save memory and facilitate efficient embedding lookups. Consequently, GlotLID-M always produces a prediction, even for n-grams that were not observed during training.

In our approach, we adopted a similar method but instead opted to store all n-grams encountered during training. This allows us to differentiate between previously seen n-grams and those unseen during training, which we classify as UNK-ngram during inference. To assess the effectiveness of this approach, we applied our model (\MS) to the \finewebtwo{} dataset, specifically focusing on instances labeled as UND\_XXX. We then apply a five-step filtering process to extract ``clean data'' from these samples:
\begin{enumerate}
    \item model's prediction label script is the same as UND\_XXX script
    \item model's prediction probability is above 95\%
    \item a document contains less than 5\% UNK-ngrams
    \item a document does not contain any space-separated letters (\eg, \textit{A B C D E ...})
    \item a document contains more than one word
\end{enumerate}

The proportion of recovered clean data using these filtering criteria is illustrated in Figure~\ref{fig:analysis_fw2_recovery}. We report results for nine scripts where script agreement exceeds 5\%. As shown, the data remains largely non-recoverable. A high level of script disagreement is likely attributable to noisy or meaningless data, where predictions from both GlotLID-M and \MS{} approach Randomness. However, we were able to recover a limited number of data points for certain scripts, including Bengali (Beng, $\sim$3\%, 1,429 samples), Gothic (Goth, $\sim$2\%, 113 samples), Hebrew (Hebr, $\sim$2\%, 4,256 samples), and Cyrillic (Cyrl, $\sim$1\%, 10,208 samples).
The recovered data primarily consists of short sentences or repetitive word patterns, yet they still align with the predicted language. For instance, the recovered Cyrillic data includes the following sentences in Russian:

\begin{itemize}
    \item \textcyrillic{ну и я ушла} (nu i ya ushla) -- \textit{well, I left too}
    \item \textcyrillic{ну и с пятницей ну и, ежели чо} (nu i s pyatnitsey nu i, yezheli cho) -- \textit{well, happy Friday, and, if anything}
    \item \textcyrillic{Народ в ЖЖ флешмобится} (Narod v ZhZh fleshmobitsya) -- \textit{people on LiveJournal are doing a flash mob.}
\end{itemize}

%% file: tables/lang_family_direction.tex
\begin{table}
    \centering
    \small
    \begin{tabular}{cc}
        \toprule
        \textbf{True → Pred} & \textbf{\# Misclassifications} \\
        \midrule
        High → High & 2182 \\
        Low → High & 528 \\
        High → Low & 440 \\
        Low → Low & 281 \\
        \bottomrule
    \end{tabular}
    \caption{Number of misclassifications by data resources (from true to predicted language resource level) in UDHR dataset using \MS}
    \label{tab:analysis_language_family_direction}
\end{table}

%% file: tables/language_misclassification.tex
\begin{table}
\centering
\begin{tabular}{l l r r}
\toprule
\textbf{$L_1$} & \textbf{$L_2$} & \textbf{\# errors} \\
\midrule
krl\_Latn & olo\_Latn & 183 (100\%) \\
bam\_Latn & dyu\_Latn & 8 (100\%) \\
hrv\_Latn & hbs\_Latn & 45 (98\%) \\
srp\_Latn & hbs\_Latn & 29 (88\%) \\
hak\_Hani & wuu\_Hani & 43 (77\%) \\
wuu\_Hani & cmn\_Hani & 3 (75\%) \\
qvn\_Latn & quy\_Latn & 23 (64\%) \\
yue\_Hani & cmn\_Hani & 8 (62\%) \\
zgh\_Tfng & tzm\_Tfng & 14 (54\%) \\
fuf\_Latn & fuv\_Latn & 8 (27\%) \\

\bottomrule
\end{tabular}
\caption{Examples of language pairs where the first language is frequently misclassified as the second.}
\label{tab:lang_misclassification}
\end{table}

%% file: figures/fineweb2_recovery_analysis.tex
\begin{figure*}[ht]
    \centering
    \includegraphics[width=1\linewidth]{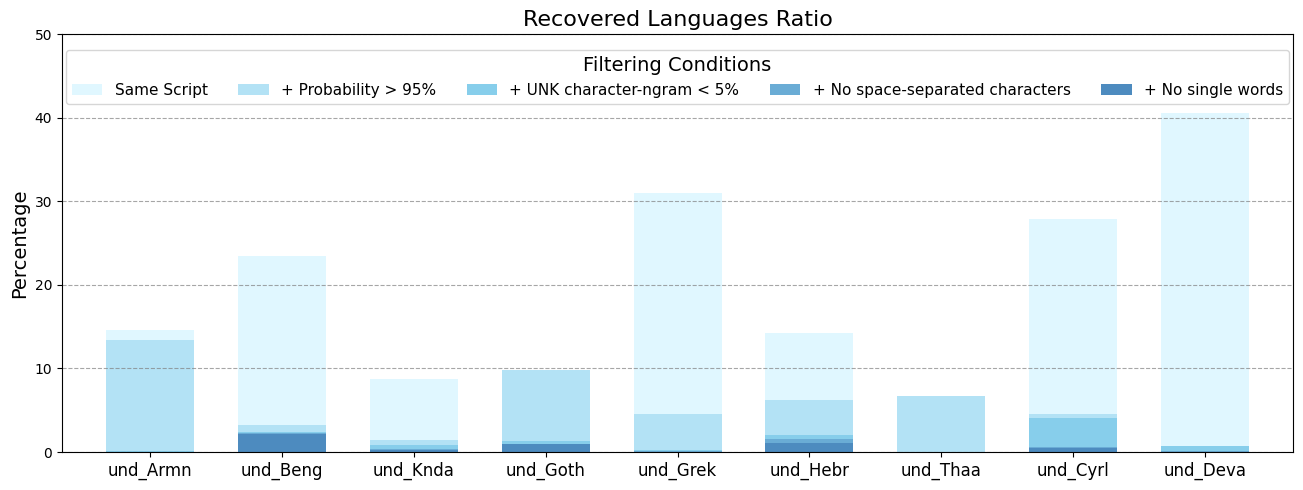}
    \caption{Ratio of the recovered text by applying 5 levels of filtering on \finewebtwo{} UND\_XXX dataset. The prediction is carried out by \MS{} model.
    }
    \label{fig:analysis_fw2_recovery}
\end{figure*}